%% file: template.tex
\documentclass[journal]{vgtc}                     

\onlineid{T-1503}

\vgtccategory{Research}

\title{SurfaceXR: Fusing Smartwatch IMUs and Egocentric Hand Pose for Seamless Surface Interactions}

\author{%
  \authororcid{Vasco Xu}{0000-0003-3990-582X}, 
  \authororcid{Brian Chen}{0009-0001-0884-3198},
  \authororcid{Eric J. Gonzalez}{0000-0002-2846-7687},
  \authororcid{Andrea Colaço}{0009-0001-6661-2216}, \and
  \authororcid{Henry Hoffmann}{0000-0003-0816-8150},
  \authororcid{Mar Gonzalez-Franco}{0000-0001-6165-4495},
  \authororcid{Karan Ahuja}{0000-0003-2497-0530}
}

\authorfooter{
  \item
  	Vasco Xu and Henry Hoffmann are with the University of Chicago. \\
  	Contact: vascoxu@uchicago.edu
  \item
  	Brian Chen is with Northwestern University. Karan Ahuja is with Northwestern University and Google. \\
  	Contact: kahuja@northwestern.edu
  \item
    The remaining co-authors are with Google. \\
    
}

\abstract{%

Mid-air gestures in Extended Reality (XR) often lead to fatigue, discomfort and imprecision, limiting their suitability for extended use. Surface-based interactions offer a compelling alternative, providing improved accuracy, speed, and comfort. However, current egocentric vision-based methods struggle with reliable surface inputs due to challenges in hand tracking and surface plane estimation from oblique and occluded viewing angles. To this extent, we introduce SurfaceXR, a novel sensor fusion approach that combines headset based hand tracking with micro-vibration data sampled from commodity smartwatch IMUs to enable precise and robust inputs on everyday surfaces. Our system is designed with flexibility in mind --- it can function using only hand tracking, only IMU sensing, or optimally with both modalities combined, and remains robust even without explicit surface calibration. Our key insight is that these modalities are complementary --- hand tracking provides 3D positional data of hand joints, whereas IMUs supply high-frequency wrist/hand motion data. Our user study across 21 participants validates SurfaceXR's effectiveness in augmenting surface touch tracking and 8 class hand-surface gesture recognition, demonstrating significant improvements over single-modality approaches. Enabled by SurfaceXR, we demonstrate a series of interactive apps for both AR and VR, ranging from on-surface sketching, text entry and gesture-based navigation. 

}


\teaser{
  \centering
  \includegraphics[width=\linewidth]{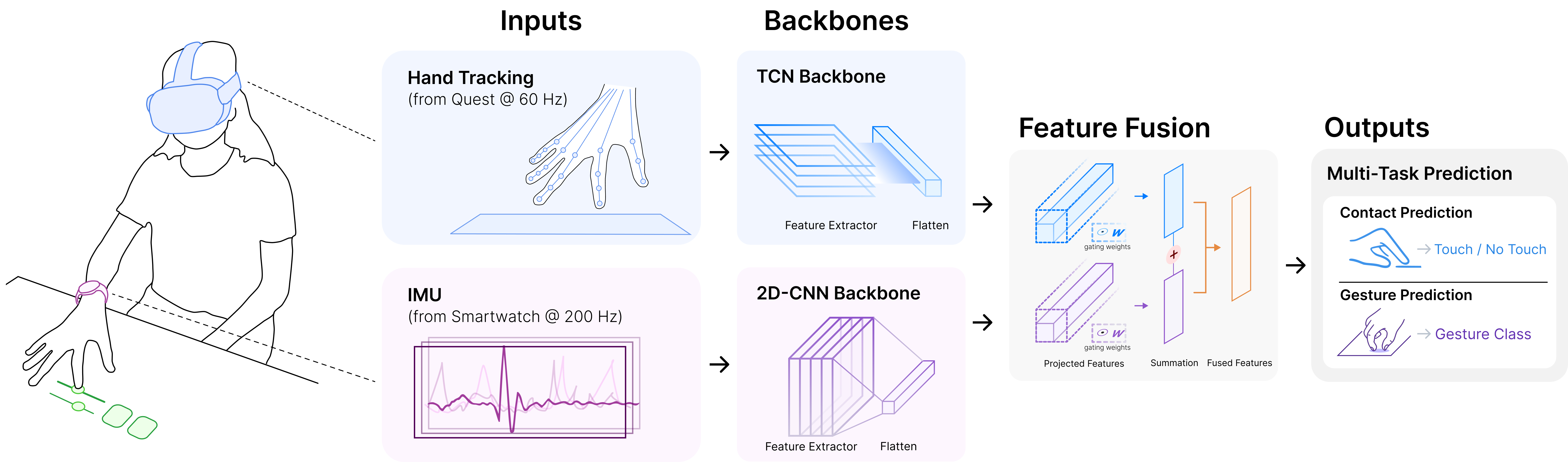}
  \caption{Overview of our system for enabling surface interactions in XR. Our network receives hand tracking data from a Quest headset and IMU data from a smartwatch. Hand keypoints are normalized relative to a surface-aligned coordinate system. A Temporal Convolutional Network (TCN) extracts features from the hand tracking data, while a 2D-CNN processes the IMU data. The features are fused using a gated fusion mechanism and used for multi-task prediction of surface contact and gesture classification. }
  \label{fig:teaser}
}

\graphicspath{{figs/}{figures/}{pictures/}{images/}{./}} 

\usepackage{tabu}                      
\usepackage{booktabs}                  
\usepackage{lipsum}                    
\usepackage{mwe}                       
\usepackage{booktabs}
\usepackage{multirow}
\usepackage[normalem]{ulem}

\usepackage{mathptmx}                  
\usepackage{amssymb}

\newcommand{\change}[1]{{\color{black}#1}}

\begin{document}


\input{sections/1_Introduction}
\input{sections/2_RelatedWork}
\input{sections/3_SurfaceXR}
\input{sections/4_Evaluation}
\input{sections/5_Applications}

\input{sections/6_Limitations}

\input{sections/7_Conclusion}

\acknowledgments{
	Vasco Xu conducted this research under Mar Gonzalez-Franco and Karan Ahuja at Google AR in Seattle, WA, USA. The authors thank the following colleagues at Google for their insightful discussions: Anish Prabhu, Ishan Chatterjee and Harish Kulkarni. Henry Hoffmann's work on this project was partly supported by the National Science Foundation (CCF-2119184 CNS-2313190 CCF-1822949 CNS-1956180). \\ \\
}

\bibliographystyle{abbrv-doi-hyperref}

\bibliography{template}








\end{document}

%% file: sections/1_Introduction.tex
\firstsection{Introduction}

\maketitle

\begin{figure}[h]
    \centering
    \includegraphics[width=\columnwidth]{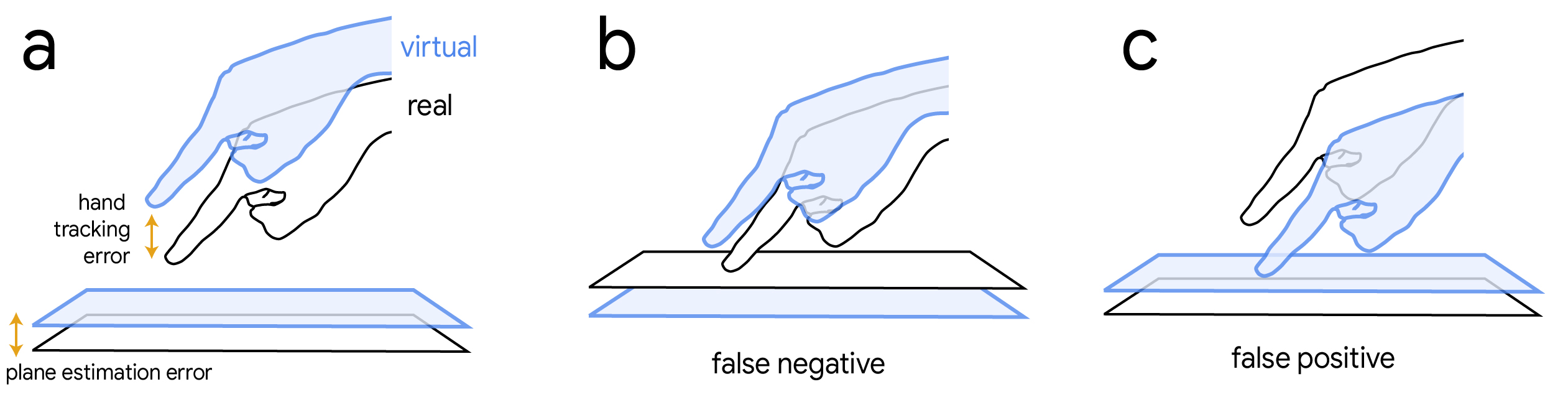}
    \caption{Accurate surface-based interactions are challenging due to errors in hand tracking and surface plane estimation. (a) There is an offset between the virtual and real hand leading to imprecise inputs: (b) False negatives, where the user is touching a surface but the system thinks it is not and (c) False positives, where the user is not touching the surface but the system thinks it is.}
    \label{fig:hand-tracking-issues}
\end{figure}

Extended Reality (XR) technologies offer immersive experiences that blend virtual and physical worlds. However, a persistent challenge in XR interaction has been the development of intuitive, precise, and comfortable input methods. While mid-air gestures have been widely adopted, they often lead to fatigue, discomfort, and imprecision during extended use, creating a ``gorilla arm" effect \cite{speicher2018selection, bowman2012questioning, armfatigue, palmeira2023quantifying, hoveroverkey} making them unsuitable for long interactive tasks. Surface-based interactions offer a compelling alternative, providing improved accuracy and comfort \cite{hansberger2017dispelling, potts2022tangibletouch}. Yet, reliably detecting and localizing surface touches in XR remains challenging, particularly for headset based egocentric camera-view methods that struggle with occlusions, oblique viewing angles, and inaccuracies in surface plane estimation \cite{dupre2024tripad} (Figure \ref{fig:hand-tracking-issues}).

To address these limitations, prior work has sought to instrument surfaces with sensors \cite{laput2019surfacesight, scratchinput, thermaltouch}. While robust, instrumenting every surface is impractical at scale. Vision-only methods, such as TriPad \cite{dupre2024tripad}, avoid external hardware using only hand tracking data with a user-registered surface but are limited by tracking and plane estimation inaccuracies. As such, other works explore complementary sensing modalities such as IMUs from rings \cite{zhao2021mouse, shen2024mousering, gu2019accurate, liang2021dualring, shadowtouch} and wrist-worn form factors \cite{meier2021tapid, streli2022taptype, gong2020acustico, kim2023vibaware, chen2023recognizing, soundscroll}. However, these approaches either require specialized hardware or are limited to only tap events, reducing their practical applicability.

In response, we propose SurfaceXR, a novel sensor fusion approach that combines headset-based hand tracking with micro-vibration data sampled from consumer smartwatch IMUs to enable precise and robust inputs on everyday surfaces. Our key insight is that these modalities provide complementary information --- hand tracking offers spatial precision (i.e., 3D positional data of hand joints) but suffers from reliability issues, while smartwatch IMUs can detect the distinctive micro-vibrations that occur during surface contact with high temporal precision. By fusing these modalities through a multimodal multi-task deep neural network (Figure \ref{fig:teaser}), SurfaceXR achieves both spatial and temporal accuracy without requiring specialized hardware or surface modifications.

A significant advantage of our approach is its flexibility across multiple operating modes: (1) hand tracking-only, which functions with just the headset's egocentric cameras; (2) IMU-only, which works even when hands are outside the camera field-of-view; and (3) multimodal mode, which provides optimal performance by dynamically weighing each modality's contribution. In our user study across 21 participants across diverse scenarios, SurfaceXR achieves a window-level touch detection F1-score of 91.2\%, cross-hair targeting error of 7.07 mm, onset latency of 47 ms, offset latency of 170 ms, and a mean surface gesture recognition F1-score of 95.0\% (single tap, double tap, swipe in four directions, pinch in/out), generalizing effectively across users. Our comprehensive evaluation includes ablation studies demonstrating the complementary nature of both sensing modalities and the system's robustness to different surface orientations and interaction contexts. SurfaceXR paves the way for improved input in various XR applications, from text entry and drawing to navigation interfaces (Figures \ref{fig:draw}, \ref{fig:trackpad}, \ref{fig:navigation}), while being especially valuable for AR scenarios where hand tracking is limited by narrow field-of-view cameras. By fusing complementary sensing modalities, our approach advances more intuitive and efficient surface-based input for XR.

%% file: sections/2_RelatedWork.tex
\section{Related Work}

SurfaceXR builds on research in on-body touch sensing using wearables, environmental touch sensing, and surface-bound interactions in XR. We situate our work at the intersection of these prior works.

On-body touch sensing has emerged as a modality for natural and quick interaction due to its constant availability and proximity to the user. Prior works \cite{harrison2010skinput, harrison2011omnitouch, egotouch} have employed wearable sensor arrays and external cameras to enable 2D touch tracking on surfaces and body. However, inaccuracies in camera based depth sensing, especially due to self-occlusion between finger and surface make touch/no-touch/hover classification challenging. To overcome such optical limitations, RF-based sensing methods such as ActiTouch \cite{actitouch} and SkinTrack \cite{zhang2016skintrack} modulate electrical signals into a wearers body to measure electrical signal flow through the body for on-skin touch detection. While these approaches offer valuable insights, they are primarily focused on on-body interactions, limiting their applicability in high throughput XR environments.

Ambient environmental sensing expands touch detection beyond a user's body and onto surrounding surfaces. Acoustic methods \cite{xiao2014toffee, paradiso2002passive, acustico} capture characteristic sounds produced when fingers tap or swipe on surfaces, but have low spatial precision and are sensitive to environmental factors such as background noise. Most accurate are optical methods such as using RGB cameras \cite{pressurevision, touchinsight, stegotype} or depth cameras \cite{xiao2013worldkit, shen2021farout, fan2022reducing}, which can capture finger-to-surface distance. Wearable devices such as rings \cite{zhao2021mouse, shen2024mousering, gu2019accurate, liang2021dualring, shadowtouch} and wristbands \cite{gong2020acustico, kim2023vibaware, chen2023recognizing, soundscroll} enable touch and gesture sensing on everyday surfaces, offering lightweight, mobile alternatives to instrumenting the environment.

Most related to our work are surface-touch interactions, which have been extensively explored as a high throughput and continuous input method for XR. Contemporary XR headsets rely on hand tracking from monochrome and depth cameras to estimate 3D hand pose. However, due to inaccuracies in hand tracking and plane estimation, detecting surface contact remains challenging. To address these limitations, researchers have explored various complementary modalities, including additional depth data \cite{xiao2013worldkit, xiao2018mrtouch, halotouch}, electrical signals \cite{actitouch}, structured laser light \cite{structuredlightspeckle} and inertial measurement units (IMUs) \cite{meier2021tapid, streli2022taptype}. However, these approaches require specialized hardware, limiting their practicality.

Recent work has made significant strides in this area. TriPad \cite{dupre2024tripad} offers a touch tracking solution that works out-of-the box on consumer headsets, using only hand tracking data with pre-registered surface planes and distance thresholds. While practical, this approach faces challenges in accurately detecting rapid or subtle touch events due to hand tracking inaccuracies and limited tracking frame rate. TapID \cite{meier2021tapid}, which is most similar to our work, addresses the limitations of hand tracking-only methods by complementing hand tracking data with IMU signals for tap detection. This sensor fusion approach mitigates issues of occlusion and tracking inaccuracies, especially during rapid finger movements. However, TapID requires specialized hardware with a high sampling rate of 1344 Hz, limiting its use and is limited to solely tap events. TapType \cite{streli2022taptype} extends TapID by incorporating Bayesian inference to improve typing in VR, but maintains the same hardware requirements.

In contrast, SurfaceXR leverages consumer IMUs found in widely available smartwatches, significantly enhancing the practicality and accessibility of our system. To overcome the challenges associated with lower sampling rates and potentially noisier data from consumer-grade IMUs, we develop a robust multimodal machine learning pipeline. This approach allows SurfaceXR to contend with the inaccuracies in hand tracking, offering a more versatile and widely applicable solution for surface touch interactions in XR environments.

%% file: sections/3_SurfaceXR.tex
\section{Dataset Capture}
To train and evaluate SurfaceXR, we collected two complementary datasets with synchronized egocentric hand tracking data from an XR headset and IMU measurements from a smartwatch. Our primary training dataset, collected from 15 participants (13 male, 2 female, mean age 25, all right-handed), captured both touch/no-touch interactions and surface gestures across horizontal (table) and vertical (wall) orientations. This comprehensive collection enables us to evaluate different factors: impact of surface plane registration, modality contributions (IMU vs. hands), and surface orientations. Our second dataset, serving as an independent test set from 6 different participants (2 male, 4 female, mean age 30, all right-handed), focused specifically on evaluating spatial touch accuracy through targeting and path tracing tasks. Together, these datasets allow us to evaluate SurfaceXR across multiple dimensions: spatial tracking accuracy, touch contact detection, temporal latency, and gesture recognition performance. \change{This study was approved by the Institutional Review Board (IRB) at Northwestern University.}

\begin{figure}[h]
    \centering
    \includegraphics[width=\columnwidth]{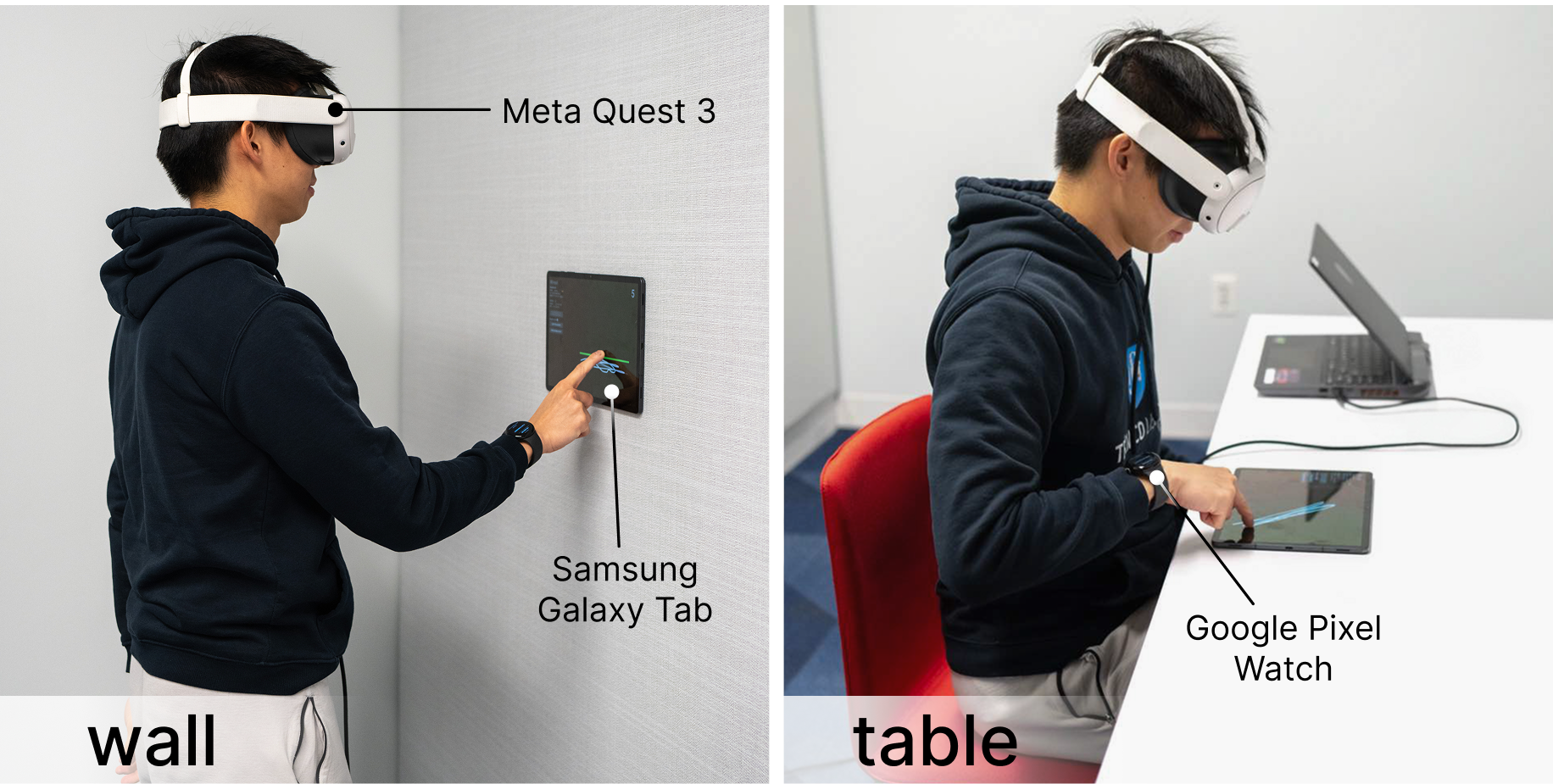}
    \caption{Our data collection setup consists of a Meta Quest 3 headset to capture hand tracking data, a Google Pixel Watch 3 to record IMU data, and Samsung Galaxy Tablet S9 to obtain ground truth annotations for surface contact and gesture events. Participants perform data collection in two conditions --- sitting at a desk and standing near a wall --- to emulate real-world interaction scenarios. }
    \label{fig:data-collection}
\end{figure}

\subsection{Apparatus}
We used the XDTK toolkit~\cite{gonzalez2024xdtk} to collect synchronized data from both the Meta Quest 3 headset's egocentric camera-based hand tracking system and a smartwatch's IMU (Google Pixel Watch 3). Ground-truth contact \change{and gesture} labels were captured from a Samsung Galaxy Tablet S9 FE (254.3 mm by 165.8 mm) touchscreen. \change{This setup allowed us to capture 34 hand-joint positions from the headset at 60 Hz, IMU data from the smartwatch at 200 Hz, and ground-truth touch events from the tablet at 60 Hz. Data from each device is transmitted at its own sampling rate and sent to Unity via the XDTK toolkit. Streams are synchronized by logging all modalities at 200 Hz using the most recent available sample from each buffer. During training, we segment data using sliding windows and remove duplicate hand tracking frames to reflect the original 60 Hz rate.}

\subsection{Gesture Data Capture}
\label{sec:gesture-protocol}
\begin{figure}[h]
    \centering
    \includegraphics[width=\columnwidth]{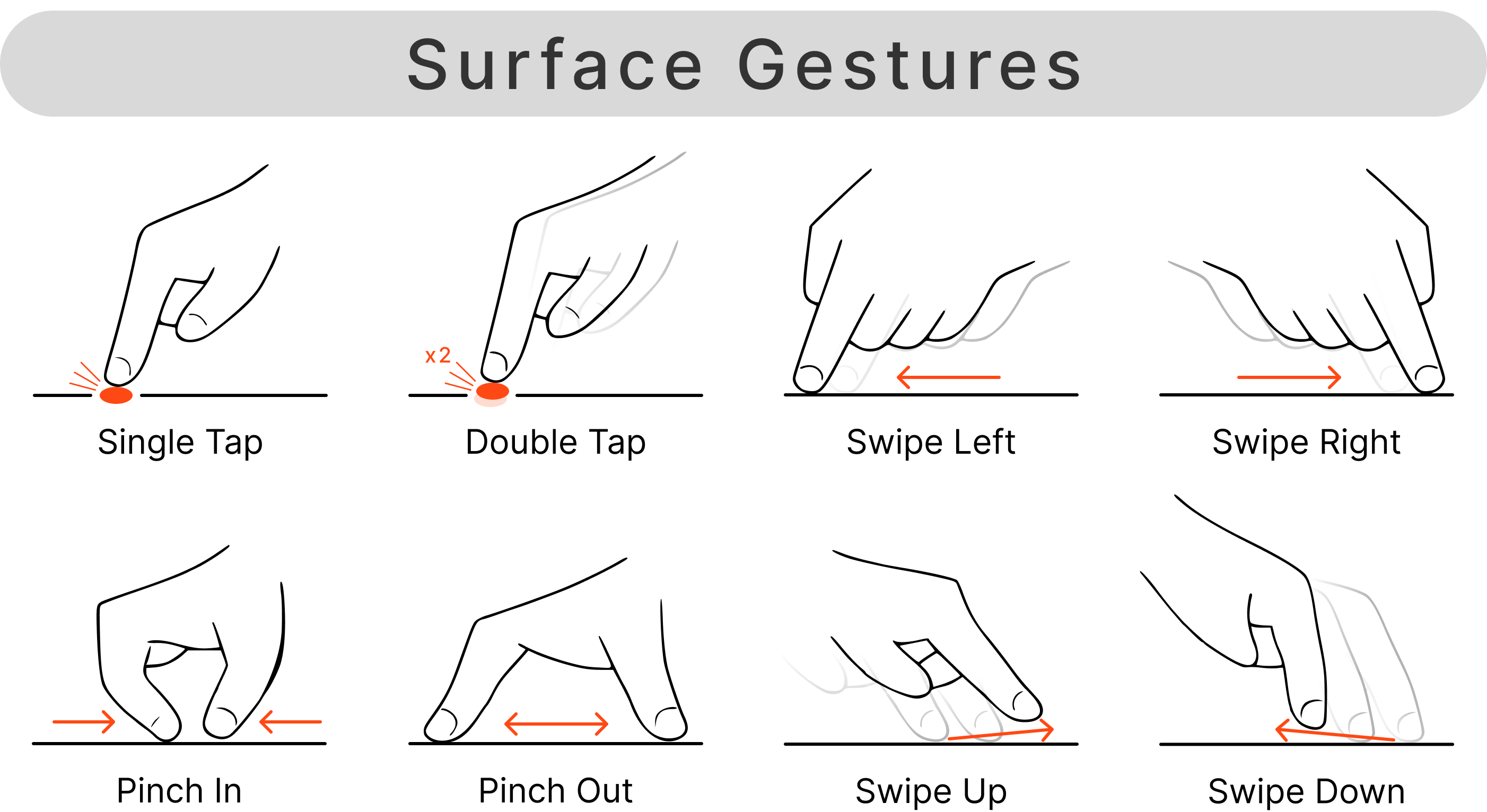}
    \caption{The SurfaceXR gesture set: single tap, double tap, swipe left, swipe right, pinch in, pinch out, swipe up and swipe down.}
    \label{fig:surface-gestures}
\end{figure}

For gesture collection, participants were asked to perform 8 common surface touch gestures inspired by the Android Touchscreen SDK \cite{android_gestures}: Single Tap, Double Tap, Swipe Left, Swipe Right, Swipe Up, Swipe Down, Pinch In, and Pinch Out (Figure \ref{fig:surface-gestures}) plus a ``Negative" class representing mid-air movements without surface contact.  We captured ground-truth data using a Samsung Galaxy Tablet, where participants were shown a random target location (indicated by a large circle) and the gesture to perform. The Android SDK's built-in touchscreen gesture recognizer provided precise timing information, with finger touch-down marking the gesture start and the recognizer identifying the gesture end. We expand this labeled interval by 30 ms before and after to capture the complete gesture motion, forming a gesture sample.

Participants repeated each gesture 20 times at randomized locations across both horizontal (table) and vertical (wall) surface orientations, yielding 4800 total gestures (15 participants $\times$ 8 gestures $\times$ 20 repetitions $\times$ 2 orientations) and 4.06 hours of gesture data. Additionally, we collected ``negative" samples where participants performed similar motions in mid-air, either as free-form movements or gesture-like actions without surface contact. In total, our gesture dataset comprises 6.52 hours of synchronized hand tracking and IMU data across all participants and sessions. For training, we extract 1-second windows from each gesture sample using a sliding window with a step size of 30 ms. Windows containing the complete gesture are assigned the corresponding label, while in-transition or idle frames are treated as negative samples.

\subsection{Touch Contact State Data Capture} 
\label{sec:touch-protocol}

Our touch contact data collection protocol follows a similar structure to that of prior work \cite{readysteadytouch}. Participants performed 6 basic movements including straight lines (``Horizontal", ``Vertical", ``Slash", and ``Backslash"), curved trajectory (``Circle") and stationary hold (``Static"). Each movement was performed 5 times for 10 seconds, followed by ``Freeform" where participants touched freely for 60 seconds to capture diverse and natural motions. 

Before each session, participants registered a virtual surface plane aligned with the physical tablet, iterating the process until they felt it was ``well-calibrated". Participants performed each touch trace under two conditions: ``Touch" (while contacting the tablet) and ``No-touch" (performed mid-air). To account for different surface orientations, touch traces were executed on both horizontal (table) and vertical (wall) surfaces. For ``Negative" no-touch data collection, participants mimicked everyday activities like washing hands, waving, using a phone, and holding objects without touching any surface. As the index finger is most common for touch interactions, the study was conducted using only the right index finger. This resulted in a total of 15 participants $\times$ 6 movements $\times$ 5 repetitions $\times$ 2 surface orientations $\times$ 2 contact states = 1800 sessions across all participants. This amounted to about 3.48 hours of touch data and 2.48 hours of negative data (no touch), creating a comprehensive dataset for training and evaluating our touch contact state detection algorithm. For training, we process each sequence using 1-second sliding windows with a 500 ms step size.

\subsection{Spatial Touch Evaluation Data Capture} 
\label{sec:spatial-protocol}

To evaluate spatial accuracy, we conducted a separate study with 6 new participants using a horizontal tablet surface as ground-truth. This independent dataset evaluated our model's generalizability beyond the training population. Following prior work on surface touch tracking \cite{xiao2018mrtouch}, participants performed two tasks:

\subsubsection{Path Tracing Task}
\label{sec:spatial-protocol-path}
Each participant traced 50 predefined shapes: 25 circles (diameters randomly varying from 5 to 10 cm) and 25 lines (in various orientations, with lengths ranging from 5 to 15 cm). A green arrow indicated the starting point, and trials automatically ended upon shape completion. For each trial, we recorded the ground-truth shape of the trace from the tablet along with synchronized hand tracking and IMU data, collecting approximately 300 traces across 6 participants.

\subsubsection{Cross-Hair Targeting Task}
\label{sec:spatial-protocol-crosshair}
Participants were instructed to touch the surface precisely at displayed cross-hairs (diameter of 2 cm), across the tablet in landscape orientation, at random locations. For each trial, we recorded the ground-truth touch position from the tablet along with synchronized hand tracking and IMU data, collecting 590
individual tap events in total across 6 participants, after removing some erroneous values.  

\section{Method}

An overview of the SurfaceXR system is shown in Figure~\ref{fig:teaser}. Our objective is to train a single multimodal multi-task neural network that can predict both surface touch contact state (touch/no-touch) and surface gesture classes from a sequence of hand tracking and IMU data. 

\subsection{Model Input}

\subsubsection{Hand Tracking Data}
\label{sec:hand_input}

Our model utilizes egocentric hand tracking data captured from the Meta Quest 3 headset at 60 Hz. While Meta records data from the entire OVR Hand Skeleton \cite{OVRHand}, our neural network processes only the index finger and thumb joints, as these are the primary fingers involved in surface gesture interactions (while touch only needs index finger). For each finger, we track 4 key joints (metacarpophalangeal, proximal interphalangeal, distal interphalangeal, and fingertip), resulting in a total of 8 tracked joints across both fingers.

To capture both positional information and dynamic movement patterns, we process two complementary data streams: (1) the 3D joint positions directly from the fingers, and (2) the joint-wise accelerations that we compute from consecutive position frames. This dual representation is particularly valuable when occlusion causes tracking errors, such as when the virtual finger appears above the surface despite actual contact (Figure \ref{fig:hand-tracking-issues}b). In these cases, acceleration data compensates by capturing characteristic contact signatures --- z-axis deceleration combined with continued x-y plane movement --- that position data alone misses.

The hand tracking input to our neural network is represented as $\mathcal{H} \in \mathbb{R}^{T \times 8 \times 3 \times 2}$, where:
\begin{itemize}
    \item T represents the temporal window (1 s = 60 samples)
    \item 8 corresponds to the per finger features (position and acceleration across 4 joints)
    \item 3 represents the spatial dimensions (x, y, z) for each joint
    \item 2 accounts for the two fingers (index and thumb)
\end{itemize}

All joint positions are normalized relative to a surface-anchored coordinate system, which may correspond to a registered physical surface (e.g., a table) or a scanned virtual interactive surface (e.g., automatic plane detection via egocentric headset SLAM \cite{unityarplane}). Specifically, hand joint positions are centered at the plane origin and rotated into surface-local coordinates, where x and y represent position along the surface and z represents distance from the surface. This normalization process minimizes differences in hand scale between users and provides consistency across different interaction scenarios and surface orientations, improving the model's generalizability. In our evaluation (Section \ref{sec:eval}), we also consider a head-based coordinate system where hands are expressed in headset coordinate space, requiring no surface plane registration. In this setup, hand positions are centered at the headset origin and rotated into head-local coordinates.

\subsubsection{IMU Inputs}

Our system captures inertial measurement data from a consumer smartwatch worn on the user's wrist. The  sensor provides 3D accelerometer and 3D gyroscope readings at a sampling rate of 200 Hz. All raw sensor values are normalized to range between -1 and 1 to ensure consistent input scaling.

To extract meaningful motion features across different frequency bands, we apply a signal processing pipeline inspired by Xu et al. \cite{xu2022enabling}. Each of the six IMU channels is processed through three Butterworth bandpass filters with frequency ranges of 0.22-8 Hz, 8-32 Hz, and 32 Hz high-pass band, implemented using causal cascaded second-order sections. This frequency decomposition preserves the original signal (as the fourth component) while isolating movements at different frequencies, allowing the model to distinguish between slow, medium, and rapid hand movements. All signal components are then detrended to mitigate noise from sensor imperfections and incidental watch motion (e.g., shifting on the wrist).

The IMU tracking input to our neural network is represented as $\mathcal{I} \in \mathbb{R}^{T \times 6 \times 4}$ where:
\begin{itemize}
    \item T represents the temporal window (1 s = 200 samples)
    \item 6 IMU channels (3 accelerometer + 3 gyroscope)
    \item 4 represents the signal components: 3 Butterworth-filtered frequency bands (0.22--8 Hz, 8--32 Hz, and 32+ Hz) and the original signal
\end{itemize}

This multi-band signal representation enables our model to detect subtle wrist dynamics during surface interactions that complement the visual hand tracking data.

\subsection{Multimodal Multi-Task Neural Network}

\subsubsection{Hand Encoder}
For processing the hand tracking input  ($\mathcal{H} \in \mathbb{R}^{2880}$), we employ a Temporal Convolutional Network (TCN) \cite{tcn} that efficiently models the sequential nature of hand movements. Inspired by STMG \cite{stmg}, our TCN comprises 10 convolutional blocks with feature dimensions (32, 32, 32, 64, 64, 64, 64, 64, 64, 64) and exponentially increasing dilation rates (1, 2, 4, 6, 8, 1, 2, 4, 6, 8) that create an expanding receptive field capturing both immediate finger movements and longer-term gestural patterns. We apply dropout ($p = 0.2$) to the temporal convolution modules to improve generalization across different users and interaction scenarios.

\subsubsection{IMU Encoder}
To extract features from inertial data ($\mathcal{I} \in \mathbb{R}^{4800}$), we adopt a 2D convolutional neural network designed to efficiently capture temporal patterns across multiple sensor channels. The network is composed of three 2D convolutional blocks, each with a convolutional layer (kernel size = $5 \times 1$) followed by group normalization, max pooling, and ReLU activation. To enhance generalization and prevent overfitting, we implement channel dropout ($p = 0.2$) during training, following the ConvBoost framework \cite{convboost}, which randomly zeros out subsets of input channels, forcing the model to learn robust, channel-independent features.

\subsubsection{Multimodal Model}
Our multimodal model uses the IMU and Hand encoder as backbones to extract features from each modality. The resulting feature embeddings are fused through a Gated Fusion mechanism \cite{gatedfusion} rather than simple feature concatenation. Each modality embedding is first projected into a shared 256-dimensional space via separate linear layers, then passed through a gating network (linear layer + sigmoid) that outputs a vector of weights between 0 and 1. These weights rescale the features according to their learned importance. The gated IMU and hand features are summed and passed through another linear layer to produce the final 256-dimensional fused representation. 

The fused representation is fed into two separate classifier heads: touch contact state detection and gesture prediction. Both classifier heads share a similar architecture, consisting of fully-connected dense layers (128 dimension, ReLU activation) with a dropout layer ($p=0.5$) followed by a softmax layer. \change{The contact state head performs binary classification (touch vs. no-touch) and produces frame-level predictions for each time step in the input sequence, enabling responsive and continuous touch state estimation. In contrast, the gesture head performs 9-class classification (8 gesture types + a no-gesture class) and predicts a single gesture label per window based on aggregated temporal context. This design allows a single model to simultaneously solve both tasks while leveraging shared multimodal representations. The resulting multimodal model contains 670k total trainable parameters (193k for hand encoder, 49k for IMU encoder, 427k for fusion and classifier layers).}

\subsection{Training Protocol}
We train the multimodal model end-to-end using the Adam optimizer \cite{adam} with a batch size of 128 and learning rate of $3 \times 10^{-4}$. We use a weighted Binary Cross Entropy Loss ($\mathcal{L}_{C}$) for training the touch contact head and standard weighted Cross Entropy ($\mathcal{L}_{G}$) for the gesture head. Loss weights are set based on the ratio of class samples. The total loss function is computed as $\mathcal{L} = \mathcal{L}_{G} + 0.5 \cdot \mathcal{L}_{C}$. The model was trained for 100 epochs at which point the model was saved for evaluation. 

Data from the first touch data collection (Section \ref{sec:touch-protocol}) also serves as negative data for gesture classification. Similarly, on-surface gesture data also serves as positive data for contact prediction. This mutual use of data across tasks provides additional training cues for both classifiers. For example, path tracing should not be recognized as gestures and surface gestures should be recognized as touches. Note that data from the independent spatial accuracy study (Section \ref{sec:spatial-protocol}) is not used for any training and is used only for evaluation.

\begin{figure}[h]
    \centering
    \includegraphics[width=\columnwidth]{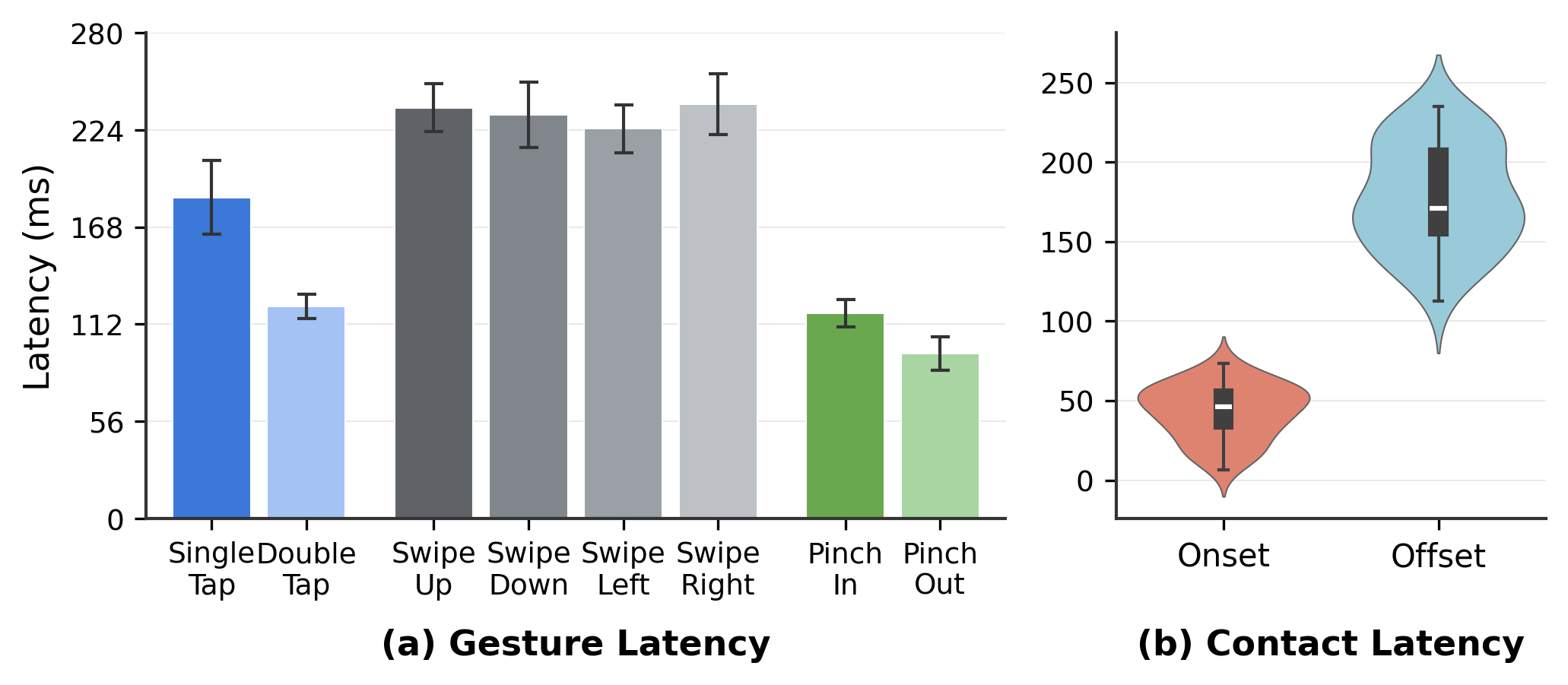}
    \caption{(a) Per-gesture recognition latency, (b) Contact state detection onset and offset latency distributions.}
    \label{fig:latency}
\end{figure}

\subsection{Real-Time Inference Pipeline}
For real-time operation, we process synchronized hand tracking and IMU streams using 1-second sliding windows (60 hand tracking samples at 60 Hz; 200 IMU samples at 200 Hz) with a stride of one frame for maximum responsiveness. To suppress transient prediction fluctuations, particularly during early gesture motion when different gestures appear similar, we apply a confidence-based hysteresis filter: state transitions require predictions to exceed 0.5 confidence for 3 consecutive frames ($\sim$50 ms at 60 Hz).

When touch contact is predicted, we determine contact coordinates (x, y) on the surface by projecting the 3D index fingertip position orthogonally onto the surface plane. When surface plane information is unavailable, we use the fingertip's (x, y) position directly.

\subsubsection{Pipeline Latency}
Our real-time pipeline streams IMU data from the Pixel Watch 3 and hand tracking data from the Quest 3 headset to a laptop (GPU: GeForce RTX 4090). IMU preprocessing (normalization and Butterworth filtering) takes 6.37 ms (SD=1.06), and hand tracking normalization takes 0.27 ms (SD=0.21). SurfaceXR's multimodal model runs inference in 6.12 ms (SD=0.69) on GPU. Postprocessing (softmax and hysteresis) adds 0.35 ms (SD=0.37), bringing the total pipeline latency to 13.61 ms (SD=1.48), enabling real-time inference at 73.5 Hz. In practice, our system runs at 60 Hz to match the hand tracking sampling rate.

\subsubsection{End-to-End Latency}
We measure end-to-end latency by counting frames in high-frequency (240 FPS) video recordings. This accounts for all communication, processing, model inference, and postprocessing. For gesture recognition (Figure \ref{fig:latency}a), we measure latency from finger lift-off until the gesture is displayed on the laptop. Pinch gestures are recognized fastest (95 - 118 ms), followed by taps (122 - 185 ms) and swipes (224 - 238 ms). Single taps are slower than double taps because the system must wait to confirm no second tap follows. Swipes are slowest because classification requires the complete finger trajectory to distinguish from other surface interactions (e.g., path tracing). For contact state detection, Figure \ref{fig:latency}b shows the distribution of onset latency (physical contact to detection) and offset latency (lift-off to detection). Our system achieves a median onset latency of 47 ms and a median offset latency of 170 ms, demonstrating responsive touch detection suitable for interactive applications. 

%% file: sections/4_Evaluation.tex
\section{Evaluation}
\label{sec:eval}
\begin{figure}[h]
    \centering
    \includegraphics[width=\columnwidth]{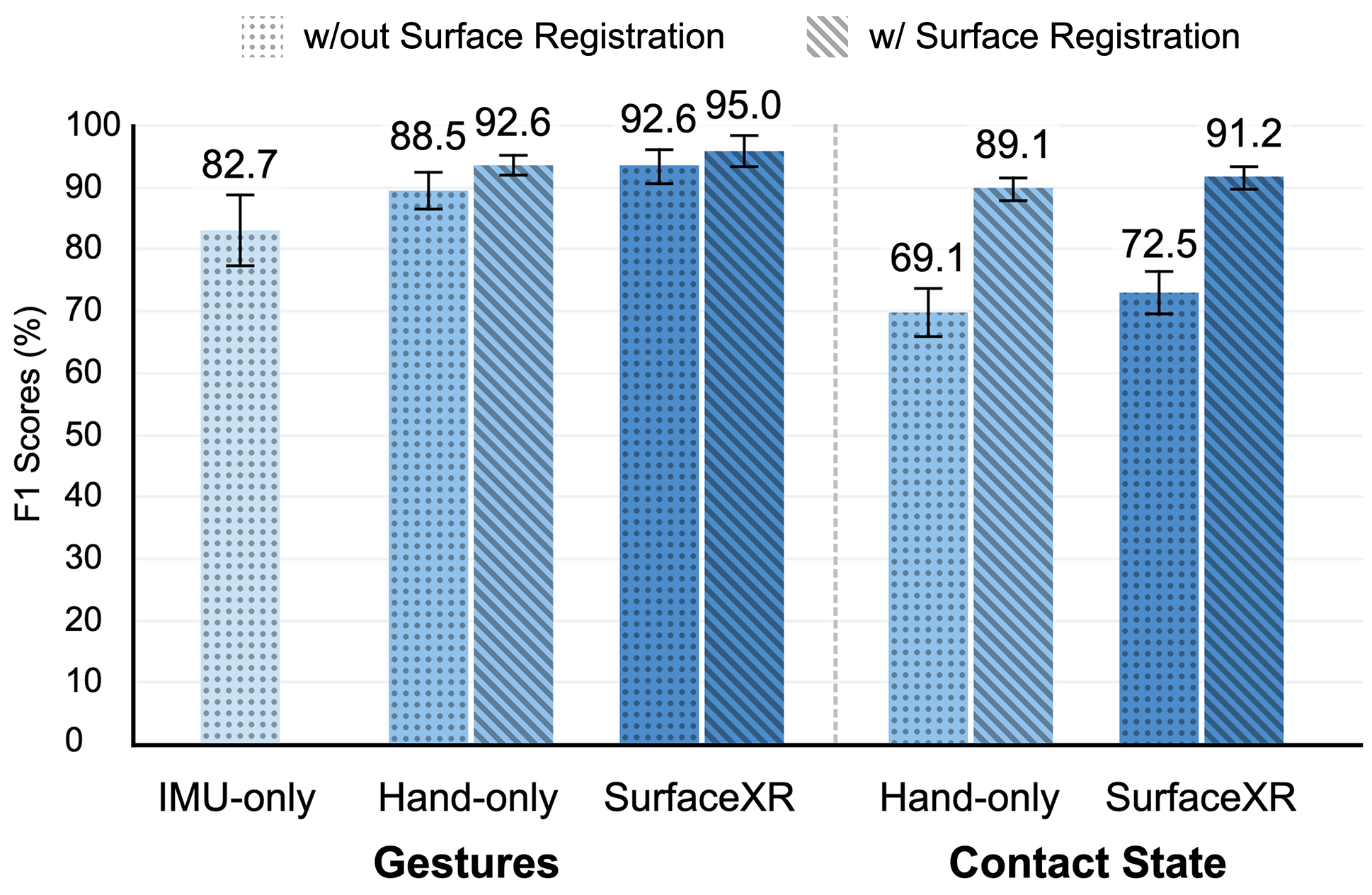}
    \caption{Summary of gesture recognition and contact state detection accuracy across different input modalities: IMU-only, Hand-only, and the multimodal model (SurfaceXR), with and without surface plane registration.}
    \label{fig:results-summary}
\end{figure}

We evaluate the accuracy of SurfaceXR across different modalities: IMU-only, hand-only, and multimodal (hand + IMU), across various tasks (contact state detection, cross-hair targeting, path tracing, gesture recognition), surface orientations and with/without surface pre-registration. For each evaluation dimension, we employ appropriate metrics to thoroughly assess our approach and quantify the impact of different factors. \change{Figure \ref{fig:results-summary} summarizes key results across modalities and conditions.} Unless otherwise noted, we refer to the multimodal system (hand + IMU) as SurfaceXR throughout our evaluation. When evaluating individual modalities, we explicitly specify IMU-only or hand-only configurations.

\subsection{Surface Gesture Recognition}

\subsubsection{Evaluation Protocol and Metrics} 
For gesture recognition, we use the dataset from Section \ref{sec:gesture-protocol} with a leave-one-participant-out cross-validation approach --- training on 14 participants and testing on the remaining holdout participant, then rotating through all 15 combinations. Performance is measured using macro F1 scores, which account for both precision and recall across all gesture classes.

\subsubsection{Overall Results} 
\begin{figure}[h]
    \centering
    \includegraphics[width=\columnwidth]{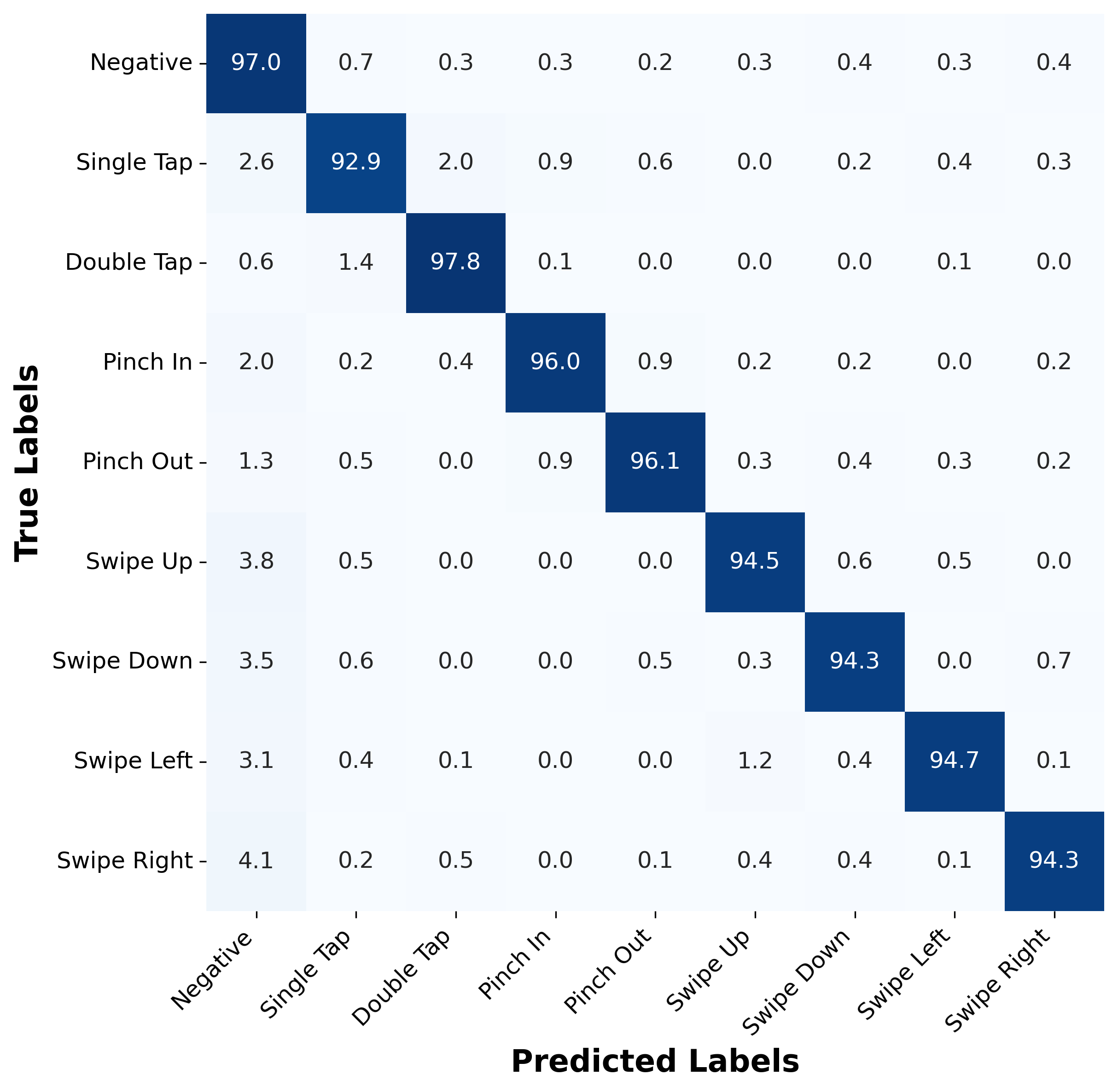}
    \caption{Normalized confusion matrix for gesture prediction (in \%) using SurfaceXR multimodal model, aggregated across all users.}
    \label{fig:gesture_conf_matrix}
\end{figure}

SurfaceXR achieves a mean F1-score of 95.0\% (precision: 95.1\%, recall: 95.2\%, SD: 2.8) across all 9 classes, and 94.7\% (precision: 94.8\%, recall: 95.0\%, SD: 3.0) when considering only the 8 gesture classes (excluding the Negative class). Figure \ref{fig:gesture_conf_matrix} presents the confusion matrix, which reveals minimal confusion between different gestures. Most classification errors occur when the model fails to detect a gesture entirely (false negatives) rather than confusing one gesture for another. Breaking down F1 scores by gesture category: taps achieve 94.0\% (precision: 93.3\%, recall: 94.9\%, SD: 4.3), swipes achieve 94.7\% (precision: 95.2\%, recall: 94.5\%, SD: 4.6), and pinches achieve 95.5\% (precision: 95.4\%, recall: 96.1\%, SD: 5.0). During data collection, users interacted with both horizontal (table) and vertical (wall) surfaces. Results show remarkable consistency across orientations: 93.4\% (SD=3.8) accuracy for horizontal surfaces versus 95.9\% (SD=2.7) for vertical surfaces.

\subsubsection{Modality Ablation} 
\label{sec:surface-modality-ablation}
When evaluating individual modalities, we found that hand-only model achieves F1=92.6\% (SD=3.8), while IMU-only performance reaches F1=82.7\% (SD=6.1). A Wilcoxon signed-rank test confirmed that the multimodal model significantly outperformed both the hand-only and IMU-only baselines ($p<.001$), with very large effect sizes ($r = 0.88$ in both cases), reflecting consistent per-participant improvements. Importantly, without surface plane registration, hand-only F1 drops by 4.4\% to 88.5\% (SD=6.4), while our multimodal approach only drops by 2.5\% to 92.6\% (SD=5.0). This demonstrates a critical advantage: SurfaceXR can operate effectively without precise surface plane registration, a common limitation in single camera XR headsets where automatic plane detection is challenging, and can even function using only IMU data when hand tracking becomes unreliable or unavailable (e.g., outside the field-of-view). Notably, IMU integration improves tap and double-tap detection by 9.4\% and 15.9\% respectively by capturing micro-vibrations from surface contact, particularly valuable when surface plane estimation is poor. 

\subsubsection{Comparison to prior works} 
To the best of our knowledge, we are the first work to explore finger-based surface gestures using consumer IMUs either standalone or with egocentric hand tracking data. Previous systems like ActualTouch \cite{readysteadytouch} achieved 90.4\% accuracy for uni-stroke gestures but required nail-mounted IMUs. Z-Ring \cite{zring} achieved 83.7\% accuracy for basic gestures (single taps, double taps, and lateral swipes) on the back of the other hand's palm using specialized finger-worn hardware. In comparison, our IMU-only model achieves 82.7\% accuracy while supporting an expanded gesture set (including vertical swipes and pinch gestures) using only commodity wrist-worn smartwatches.

\subsubsection{Tap Event Detection}
Most prior works have focused on tap detection as a task that generalizes to various XR scenarios such as UI selection, typing, among many others \cite{xiao2018mrtouch, meier2021tapid,streli2022taptype}. Using the IMU-only model, our single tap detection (F1=80.7\%) is significantly lower than TapID (F1=99.7\%) by roughly 19.0\%, which can be attributed to (a) our significantly lower sampling rate (200 Hz vs. 1344 Hz) and (b) the substantial confusion between single and double taps. However, when we consider single taps and double taps as a unified ``tap" class, this improves considerably from 80.7\% to 88.6\%. Note, there are still 6 other on surface gesture classes that lead to ambiguity in predictions. Our multimodal model achieves stronger tap detection: F1=92.0\% for single taps and F1=97.0\% for double taps. Using the combined tap class, multimodal achieves F1=96.3\% with surface plane registration, dropping only to 95.5\% without, demonstrating that IMU integration provides resilience when surface plane estimation is unavailable. This compares favorably to MRTouch (96.5\%), which requires an additional depth camera. In comparison, the hand-only model achieves 94.5\% with surface plane registration, dropping to 88.8\% without. 



\subsection{Contact State Detection}

\subsubsection{Evaluation Protocol and Metrics} 
For contact state detection, we use the dataset from Section \ref{sec:touch-protocol} with a leave-one-participant-out cross-validation approach --- training on 14 participants and testing on the remaining holdout participant, then rotating through all 15 combinations. We report macro F1 scores for window-level classification accuracy.

\subsubsection{Overall Results} 
SurfaceXR achieves an average F1-score of 91.2\% (precision: 91.2\%, recall: 91.4\%, SD: 4.2) for window-level contact state detection across all participants. The balanced precision and recall indicate that our system reliably detects touch events while limiting false positives, even in challenging scenarios with noisy surface plane registration and negative samples where users hover closely above surfaces. Performance is consistent across surface orientations, with F1-scores of 89.3\% (SD=7.2) for vertical surfaces and 92.5\% (SD=3.9) for horizontal surfaces.

\subsubsection{Comparison with Hand-Surface Threshold Baseline} 
Our hand-only model achieves a mean contact state F1-score of 89.1\% (vs 91.2\% for our multimodal model). To establish a baseline for comparison, we implemented a hand-to-surface threshold based model akin to Tripad \cite{dupre2024tripad}. For each test user, we compute the threshold as the average index finger to surface distance from contact points across all train users. This baseline achieved a surface contact F1-score of 74.0\% (precision: 90.8\%, recall: 62.5\%, SD: 7.5). High precision indicates that most detected contact points are reliable, but low recall informs us that many contact points are missed, likely due to errors in hand tracking and inaccurate surface plane registration. This relatively low performance for the baseline technique reflects the challenging nature of our dataset, which includes noisy surface plane registrations and close hover scenarios that confound distance-based approaches. Paired $t$-tests confirm that the multimodal model significantly outperforms both the threshold-based baseline ($p<.001$, $d=1.74$) and the hand-only model ($p=.0025$, $d=1.04$), with large effect sizes. Importantly, the improvement over hand-only tracking demonstrates that IMU sensing provides complementary contact cues beyond vision alone. Collectively, these results highlight the limitations of using threshold-based techniques for accurate touch tracking in XR.

\subsection{Path Tracing} 

\subsubsection{Evaluation Protocol and Metrics} 
We evaluated path tracing performance using our independent test dataset of 6 participants (Section \ref{sec:spatial-protocol-path}) and training a model on the first dataset (Section \ref{sec:gesture-protocol} and \ref{sec:touch-protocol}). Participants traced predefined shapes displayed on a touchscreen tablet, which recorded ground-truth touch positions. The touchscreen provided visual feedback of the actual traces, but participants received no feedback from SurfaceXR predictions. For each traced shape, we computed the absolute Euclidean distance between the fingertip position on the surface and the nearest point along the shape, following MRTouch \cite{xiao2018mrtouch}.

\subsubsection{Overall Results} 
Across all participants and shapes, we measured a mean user tracing error of 12.4 mm. We observed that circle tracing resulted in slightly higher prediction errors (15.4 mm) compared to straight line tracing (9.53 mm), likely due to the more complex continuous curvature requiring finer motor control. We note that there is inherent error when projecting the 3D fingertips onto the virtual surface, which contributes to spatial error.

\subsubsection{Comparison to prior works}
SurfaceXR's spatial error of 12.4 mm demonstrates reasonable performance, though precise touch localization is not our primary goal. Unlike depth camera-based specialized systems like MRTouch (5.4 mm) \cite{xiao2018mrtouch} and OmniTouch (11.7 mm) \cite{harrison2011omnitouch} that focus exclusively on spatial accuracy through dedicated hardware or calibration requirements, we deliberately use Quest's built-in hand tracking without modification to prioritize accessibility and deployability. These results should be viewed as a baseline that indicates promising avenues for future exploration, such as directly regressing on image data or incorporating additional visual modalities. Note our current approach offers significant advantages in hardware simplicity and calibration-free operation that will inherently benefit from ongoing improvements in consumer XR hand tracking technology.

\subsection{Cross-hair Targeting}

\subsubsection{Evaluation Protocol and Metrics} 
For this task, we trained a model on all 15 participants from our first dataset (Section \ref{sec:gesture-protocol} and \ref{sec:touch-protocol}) and evaluated it on the 6 participants from our independent test dataset (Section \ref{sec:spatial-protocol-crosshair}). We measured performance using spatial accuracy quantified by the Euclidean distance between the target cross-hair and the detected touch point (in centimeters).

\subsubsection{Overall Results} SurfaceXR achieved a mean targeting error of 7.07 mm (SD=3.9) across all participants and conditions. In spatial accuracy, SurfaceXR's tap mean error compares well with MRTouch's reported accuracy of 5.4 mm, without requiring additional depth cameras. Note, that unlike MRTouch, which removed about 2\% of their data as outliers, we did not perform any outlier removal. 

%% file: sections/5_Applications.tex
\section{Applications}
\begin{figure}[h]
    \centering
    \includegraphics[width=\columnwidth]{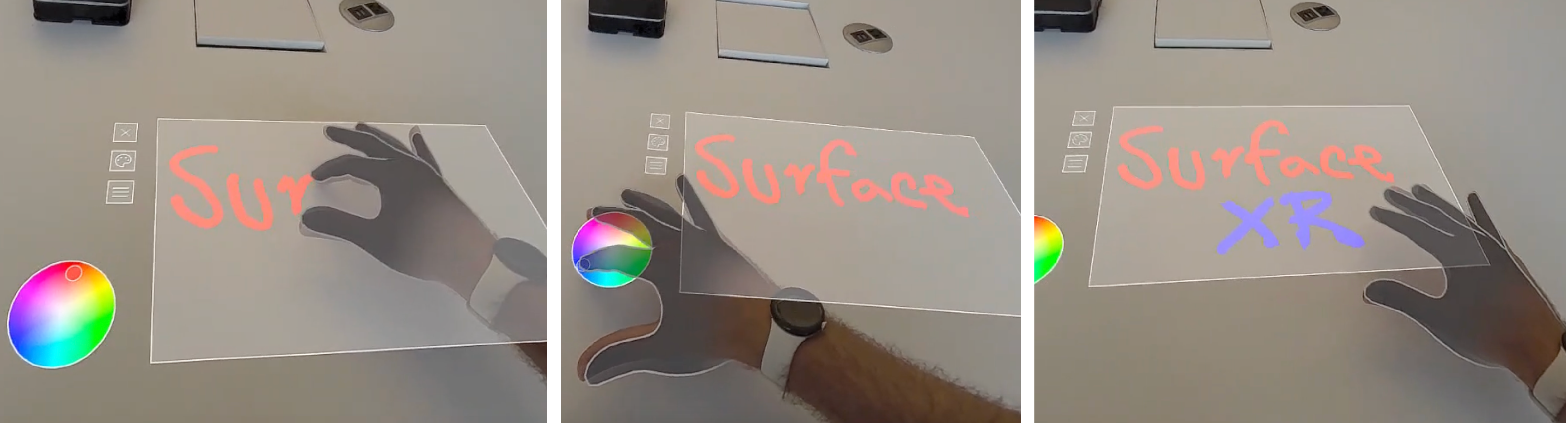}
    \caption{Drawing application in XR with color selection and tap to clear canvas.}
    \label{fig:draw}
\end{figure}

We implemented several interactive applications for both AR and XR to showcase the potential of SurfaceXR. We develop a drawing application that uses precise surface contact detection to enable a natural sketching experience (\autoref{fig:draw}). Users can select colors from a virtual palette by simply touching the surface and can adjust the stroke width by altering their finger angle. By anchoring the canvas to a surface, we can provide greater accuracy with less fatigue \cite{cheng2022comfortable}.

\begin{figure}[h]
    \centering
    \includegraphics[width=\columnwidth]{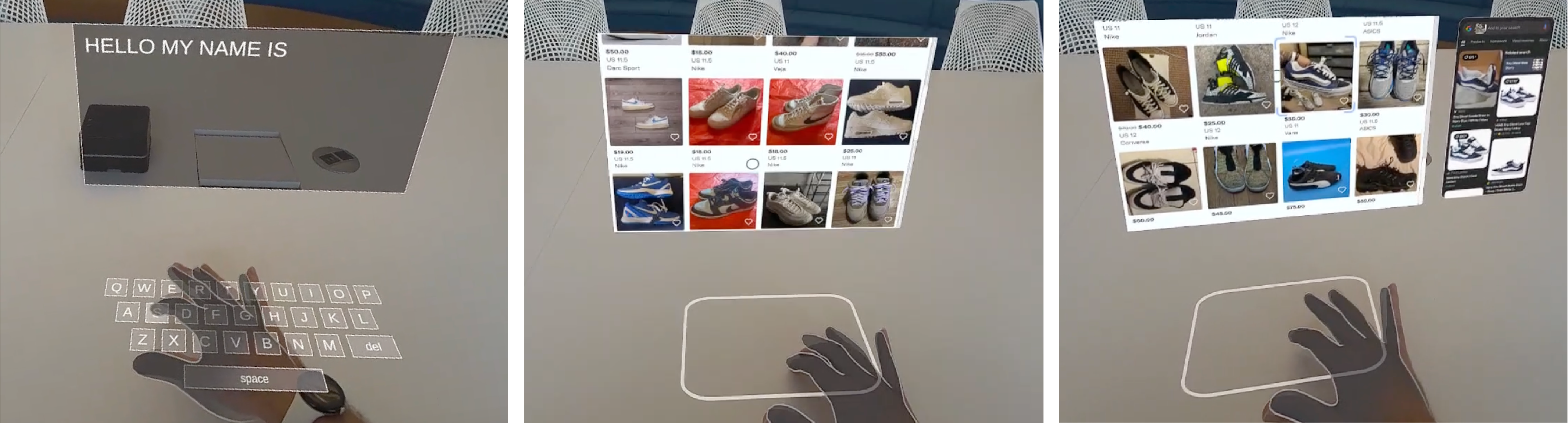}
    \caption{The system offers enough precision for text entry while its comprehensive gesture detection functions as a complete trackpad --- providing cursor control, scroll swipes, and double tap selection.}
    \label{fig:trackpad}
\end{figure}

We further developed a dual-input interface that combines a virtual keyboard with a trackpad, both anchored to a table (\autoref{fig:trackpad}). In this setup, users can type on the virtual keyboard and use the trackpad to navigate documents and content --- emulating the familiar experience of a computer trackpad. This integrated system offers a more natural and precise interaction surface compared to standard hand-ray input. Furthermore, when hand tracking becomes unreliable due to occlusion, the system seamlessly falls back to relying solely on the smartwatch IMU.

\begin{figure}[h]
    \centering
    \includegraphics[width=\columnwidth]{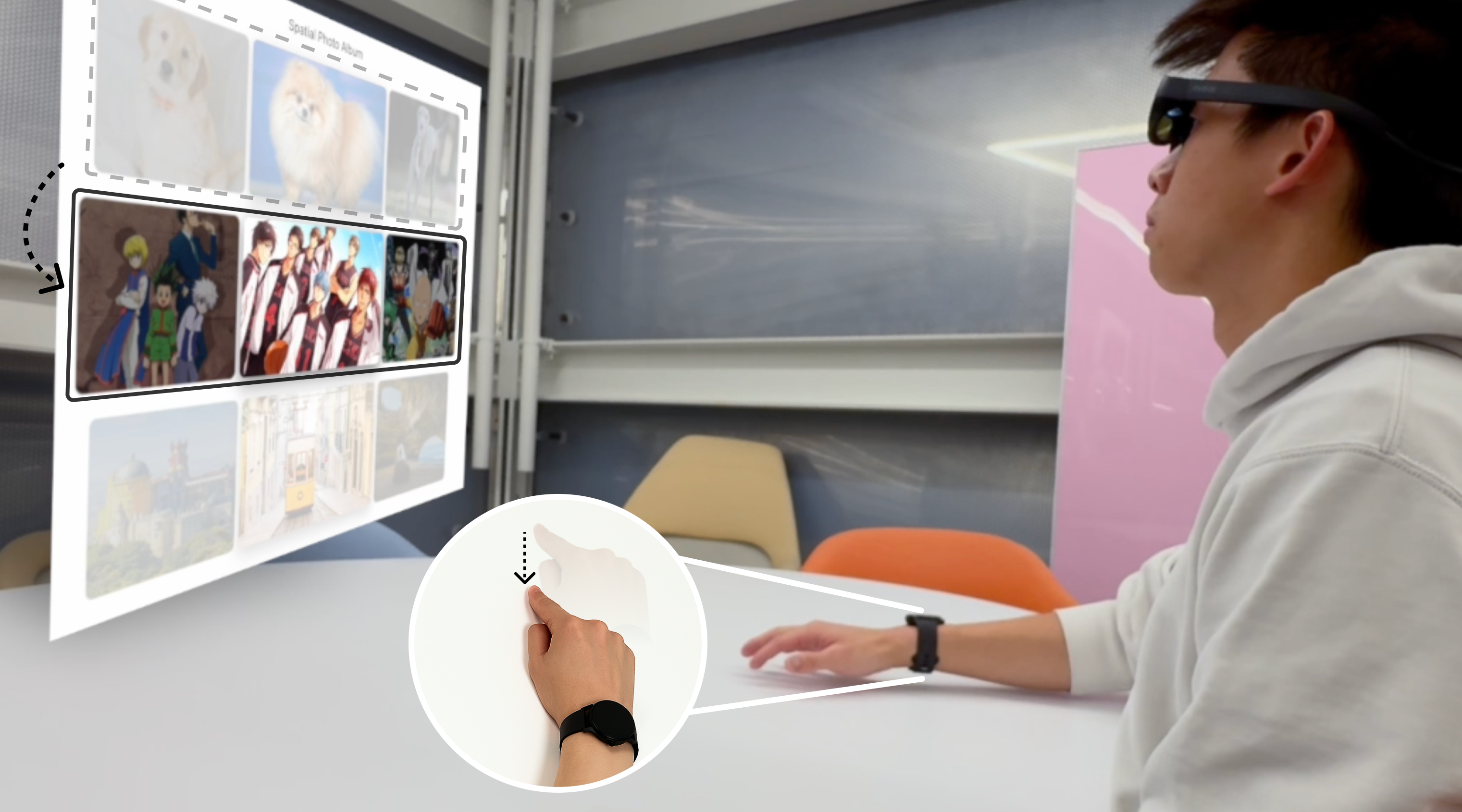}
    \caption{IMU alone can be used to support AR application on glasses.}
    \label{fig:navigation}
\end{figure}

We also designed a interactive photo album interface for AR glasses (i.e., XReal \cite{xreal2023}). This application uses only the smartwatch IMU since the hands are typically out-of-view during surface interactions \cite{hoveroverkey,gonzalez2024intent}. Users swipe to navigate through the photos, single tap to select items, and double tap to deselect. Pinch gestures are used to zoom in and out, offering an intuitive and familiar interaction model for navigating virtual content in AR (\autoref{fig:navigation}).

%% file: sections/6_Limitations.tex
\section{Limitations and Future Work}

While SurfaceXR demonstrates significant advancements in surface touch interaction for XR, several limitations remain. Our evaluation focused solely on single-touch events using the index finger, limiting our understanding of multi-touch performance. The system is also restricted to interactions with the hand wearing the smartwatch, limiting bi-manual input such as two-handed typing. Additionally, the smartwatch itself may subtly influence interaction behavior, as users may tap or gesture differently compared with an uninstrumented hand. Moreover, extreme occlusion cases where the hand is completely obscured from the headset's view can still pose challenges, despite our multimodal approach. In such scenarios, exploring the integration of additional sensing modalities available on modern smartwatches, such as audio, Ultra-Wideband (UWB), or RF sensing, could potentially improve accuracy and robustness. Future work could also integrate depth maps, which are becoming increasingly available to developers on modern headsets to improve reliability of touch tracking and surface plane registration. 

SurfaceXR has currently only been tested on flat, rigid surfaces. While the tap impulse should manifest on surfaces, regardless of their contour, future iterations still need to evaluate the efficacy of SurfaceXR in diverse environments, including non-planar and soft surfaces. Additionally, our participant pool consisted entirely of right-handed individuals with a mean age of 25, and our evaluation focused primarily on technical performance. Future studies should explore broader populations and usability factors such as learning curves, user preferences, social acceptability, and ergonomic impacts during extended use of surface-based interactions.

Another key area for future research is optimizing SurfaceXR for power efficiency to enable on-device deployment. This could involve developing more efficient machine learning models, exploring intermittent sensing strategies, or leveraging low-power neural processors found in many smartwatches. Addressing these challenges and exploring these directions will further enhance SurfaceXR's capabilities and its potential impact on XR interaction paradigms.

%% file: sections/7_Conclusion.tex
\section{Conclusion}

In this paper, we introduced SurfaceXR, a novel sensor fusion approach that combines headset-based hand tracking with micro-vibration data sensed from commodity smartwatch IMUs to enable precise and robust surface interactions on everyday surfaces, overcoming the limitations of traditional mid-air gestures and current vision-based methods. Our user study across 21 participants demonstrated SurfaceXR's effectiveness, achieving a window-level touch detection F1-score of 91.2\% and a gesture recognition F1-score of 95.0\%, generalizing across users and surface orientations. Importantly, SurfaceXR achieves these results using off-the-shelf hardware, enhancing its potential for widespread adoption. These results validate SurfaceXR's potential to enable natural and comfortable surface interactions in XR, paving the way for improved input in various applications such as text entry, sketching, and other immersive tasks.

%% file: template.bib
@online{android_gestures,
  author={{Android Developers}},
  title={Detect common gestures},
  year={2024},
  url={https://developer.android.com/develop/ui/views/touch-and-input/gestures/detector},
  urldate={2025-03-26}
}

@article{adam,
  title={Adam: A method for stochastic optimization},
  author={Kingma, Diederik P and Ba, Jimmy},
  journal={arXiv preprint arXiv:1412.6980},
  year={2014}
}

@article{gatedfusion,
  title={Gated multimodal units for information fusion},
  author={Arevalo, John and Solorio, Thamar and Montes-y-G{\'o}mez, Manuel and Gonz{\'a}lez, Fabio A},
  journal={arXiv preprint arXiv:1702.01992},
  year={2017}
}

@inproceedings{cheng2022comfortable,
  title={ComforTable user interfaces: Surfaces reduce input error, time, and exertion for tabletop and mid-air user interfaces},
  author={Cheng, Yi Fei and Luong, Tiffany and Fender, Andreas Rene and Streli, Paul and Holz, Christian},
  booktitle={2022 IEEE International Symposium on Mixed and Augmented Reality (ISMAR)},
  pages={150--159},
  year={2022},
  organization={IEEE}
}

@inproceedings{gonzalez2024xdtk,
  title={Xdtk: A cross-device toolkit for input \& interaction in xr},
  author={Gonzalez, Eric J and Patel, Khushman and Ahuja, Karan and Gonzalez-Franco, Mar},
  booktitle={2024 IEEE Conference on Virtual Reality and 3D User Interfaces Abstracts and Workshops (VRW)},
  pages={467--470},
  year={2024},
  organization={IEEE}
}

@inproceedings{gonzalez2024intent,
  title={Intent-driven input device arbitration for XR},
  author={Gonzalez, Eric J and Chatterjee, Ishan and Gonzalez-Franco, Mar and Cola{\c{c}}o, Andrea and Ahuja, Karan},
  booktitle={Extended Abstracts of the CHI Conference on Human Factors in Computing Systems},
  pages={1--5},
  year={2024}
}

@inproceedings{tcn,
  title={Temporal convolutional networks for action segmentation and detection},
  author={Lea, Colin and Flynn, Michael D and Vidal, Rene and Reiter, Austin and Hager, Gregory D},
  booktitle={proceedings of the IEEE Conference on Computer Vision and Pattern Recognition},
  pages={156--165},
  year={2017}
}

@article{convboost,
  title={ConvBoost: Boosting ConvNets for sensor-based activity recognition},
  author={Shao, Shuai and Guan, Yu and Zhai, Bing and Missier, Paolo and Pl{\"o}tz, Thomas},
  journal={Proceedings of the ACM on Interactive, Mobile, Wearable and Ubiquitous Technologies},
  volume={7},
  number={2},
  pages={1--21},
  year={2023},
  publisher={ACM New York, NY, USA}
}

@inproceedings{stmg,
  title={STMG: A Machine Learning Microgesture Recognition System for Supporting Thumb-Based VR/AR Input},
  author={Kin, Kenrick and Wan, Chengde and Koh, Ken and Marin, Andrei and Camg{\"o}z, Necati Cihan and Zhang, Yubo and Cai, Yujun and Kovalev, Fedor and Ben-Zacharia, Moshe and Hoople, Shannon and others},
  booktitle={Proceedings of the 2024 CHI Conference on Human Factors in Computing Systems},
  pages={1--15},
  year={2024}
}

@article{readysteadytouch,
  title={Ready, steady, touch! sensing physical contact with a finger-mounted imu},
  author={Shi, Yilei and Zhang, Haimo and Zhao, Kaixing and Cao, Jiashuo and Sun, Mengmeng and Nanayakkara, Suranga},
  journal={Proceedings of the ACM on Interactive, Mobile, Wearable and Ubiquitous Technologies},
  volume={4},
  number={2},
  pages={1--25},
  year={2020},
  publisher={ACM New York, NY, USA}
}

@inproceedings{chen2023recognizing,
  title={Recognizing On-Surface Gesture Using Smartwatch},
  author={Chen, Junyu and Saito, Hiroshi and Nakamura, Hiroshi},
  booktitle={IEICE Conferences Archives},
  year={2023},
  organization={The Institute of Electronics, Information and Communication Engineers}
}

@inproceedings{stegotype,
  title={Stegotype: Surface typing from egocentric cameras},
  author={Richardson, Mark and Botros, Fadi and Shi, Yangyang and Guo, Pinhao and Snow, Bradford J and Zhang, Linguang and Dong, Jingming and Vertanen, Keith and Ma, Shugao and Wang, Robert},
  booktitle={Proceedings of the 37th Annual ACM Symposium on User Interface Software and Technology},
  pages={1--14},
  year={2024}
}

@inproceedings{armfatigue,
  title={Modeling cumulative arm fatigue in mid-air interaction based on perceived exertion and kinetics of arm motion},
  author={Jang, Sujin and Stuerzlinger, Wolfgang and Ambike, Satyajit and Ramani, Karthik},
  booktitle={Proceedings of the 2017 CHI conference on human factors in computing systems},
  pages={3328--3339},
  year={2017}
}

@inproceedings{soundscroll,
  title={SoundScroll: Robust Finger Slide Detection Using Friction Sound and Wrist-Worn Microphones},
  author={Kim, Daehwa and Whitmire, Eric and Boldu, Roger and Kienzle, Wolf and Benko, Hrvoje},
  booktitle={Proceedings of the 2024 ACM International Symposium on Wearable Computers},
  pages={63--70},
  year={2024}
}

@inproceedings{egotouch,
  title={EgoTouch: On-Body Touch Input Using AR/VR Headset Cameras},
  author={Mollyn, Vimal and Harrison, Chris},
  booktitle={Proceedings of the 37th Annual ACM Symposium on User Interface Software and Technology},
  pages={1--11},
  year={2024}
}

@inproceedings{hoveroverkey,
  title={Hovering Over the Key to Text Input in XR},
  author={Gonzalez-Franco, Mar and Abdlkarim, Diar and Bhatia, Arpit and Macgregor, Stuart and Fotso-Puepi, Jason A and Gonzalez, Eric J and Seifi, Hasti and Di Luca, Massimiliano and Ahuja, Karan},
  booktitle={2024 IEEE International Symposium on Emerging Metaverse (ISEMV)},
  pages={13--16},
  year={2024},
  organization={IEEE}
}

@inproceedings{actitouch,
  title={ActiTouch: Robust touch detection for on-skin AR/VR interfaces},
  author={Zhang, Yang and Kienzle, Wolf and Ma, Yanjun and Ng, Shiu S and Benko, Hrvoje and Harrison, Chris},
  booktitle={Proceedings of the 32nd Annual ACM Symposium on User Interface Software and Technology},
  pages={1151--1159},
  year={2019}
}

@inproceedings{meier2021tapid,
  title={TapID: Rapid touch interaction in virtual reality using wearable sensing},
  author={Meier, Manuel and Streli, Paul and Fender, Andreas and Holz, Christian},
  booktitle={2021 IEEE Virtual Reality and 3D User Interfaces (VR)},
  pages={519--528},
  year={2021},
  organization={IEEE}
}

@inproceedings{dupre2024tripad,
  title={TriPad: Touch Input in AR on Ordinary Surfaces with Hand Tracking Only},
  author={Dupr{\'e}, Camille and Appert, Caroline and Rey, St{\'e}phanie and Saidi, Houssem and Pietriga, Emmanuel},
  booktitle={Proceedings of the CHI Conference on Human Factors in Computing Systems},
  pages={1--18},
  year={2024}
}

@inproceedings{harrison2010skinput,
  title={Skinput: appropriating the body as an input surface},
  author={Harrison, Chris and Tan, Desney and Morris, Dan},
  booktitle={Proceedings of the SIGCHI conference on human factors in computing systems},
  pages={453--462},
  year={2010}
}

@inproceedings{harrison2011omnitouch,
  title={OmniTouch: wearable multitouch interaction everywhere},
  author={Harrison, Chris and Benko, Hrvoje and Wilson, Andrew D},
  booktitle={Proceedings of the 24th annual ACM symposium on User interface software and technology},
  pages={441--450},
  year={2011}
}

@inproceedings{gu2019accurate,
  title={Accurate and low-latency sensing of touch contact on any surface with finger-worn IMU sensor},
  author={Gu, Yizheng and Yu, Chun and Li, Zhipeng and Li, Weiqi and Xu, Shuchang and Wei, Xiaoying and Shi, Yuanchun},
  booktitle={Proceedings of the 32nd annual ACM symposium on user interface software and technology},
  pages={1059--1070},
  year={2019}
}

@article{xiao2018mrtouch,
  title={MRTouch: Adding touch input to head-mounted mixed reality},
  author={Xiao, Robert and Schwarz, Julia and Throm, Nick and Wilson, Andrew D and Benko, Hrvoje},
  journal={IEEE transactions on visualization and computer graphics},
  volume={24},
  number={4},
  pages={1653--1660},
  year={2018},
  publisher={IEEE}
}

@inproceedings{xiao2013worldkit,
  title={WorldKit: rapid and easy creation of ad-hoc interactive applications on everyday surfaces},
  author={Xiao, Robert and Harrison, Chris and Hudson, Scott E},
  booktitle={Proceedings of the SIGCHI Conference on Human Factors in Computing Systems},
  pages={879--888},
  year={2013}
}

@inproceedings{shen2021farout,
  title={Farout touch: Extending the range of ad hoc touch sensing with depth cameras},
  author={Shen, Vivian and Spann, James and Harrison, Chris},
  booktitle={Proceedings of the 2021 ACM Symposium on Spatial User Interaction},
  pages={1--12},
  year={2021}
}

@inproceedings{laput2019surfacesight,
  title={SurfaceSight: a new spin on touch, user, and object sensing for IoT experiences},
  author={Laput, Gierad and Harrison, Chris},
  booktitle={Proceedings of the 2019 CHI Conference on Human Factors in Computing Systems},
  pages={1--12},
  year={2019}
}

@inproceedings{xiao2014toffee,
  title={Toffee: enabling ad hoc, around-device interaction with acoustic time-of-arrival correlation},
  author={Xiao, Robert and Lew, Greg and Marsanico, James and Hariharan, Divya and Hudson, Scott and Harrison, Chris},
  booktitle={Proceedings of the 16th international conference on Human-computer interaction with mobile devices \& services},
  pages={67--76},
  year={2014}
}

@inproceedings{paradiso2002passive,
  title={Passive acoustic sensing for tracking knocks atop large interactive displays},
  author={Paradiso, Joseph A and Leo, Che King and Checka, Nisha and Hsiao, Kaijen},
  booktitle={SENSORS, 2002 IEEE},
  volume={1},
  pages={521--527},
  year={2002},
  organization={IEEE}
}

@inproceedings{zhang2016skintrack,
  title={Skintrack: Using the body as an electrical waveguide for continuous finger tracking on the skin},
  author={Zhang, Yang and Zhou, Junhan and Laput, Gierad and Harrison, Chris},
  booktitle={Proceedings of the 2016 CHI Conference on Human Factors in Computing Systems},
  pages={1491--1503},
  year={2016}
}

@article{zhao2021mouse,
  title={Mouse on a Ring: A Mouse Action Scheme Based on IMU and Multi-Level Decision Algorithm},
  author={Zhao, Yuliang and Ren, Xianshou and Lian, Chao and Han, Kunyu and Xin, Liming and Li, Wen J},
  journal={IEEE Sensors Journal},
  volume={21},
  number={18},
  pages={20512--20520},
  year={2021},
  publisher={IEEE}
}

@inproceedings{shen2024mousering,
  title={MouseRing: Always-available Touchpad Interaction with IMU Rings},
  author={Shen, Xiyuan and Yu, Chun and Wang, Xutong and Liang, Chen and Chen, Haozhan and Shi, Yuanchun},
  booktitle={Proceedings of the CHI Conference on Human Factors in Computing Systems},
  pages={1--19},
  year={2024}
}

@inproceedings{gong2020acustico,
  title={Acustico: Surface tap detection and localization using wrist-based acoustic TDOA sensing},
  author={Gong, Jun and Gupta, Aakar and Benko, Hrvoje},
  booktitle={Proceedings of the 33rd annual acm symposium on user interface software and technology},
  pages={406--419},
  year={2020}
}

@article{liang2021dualring,
  title={DualRing: Enabling subtle and expressive hand interaction with dual IMU rings},
  author={Liang, Chen and Yu, Chun and Qin, Yue and Wang, Yuntao and Shi, Yuanchun},
  journal={Proceedings of the ACM on Interactive, Mobile, Wearable and Ubiquitous Technologies},
  volume={5},
  number={3},
  pages={1--27},
  year={2021},
  publisher={ACM New York, NY, USA}
}

@inproceedings{kim2023vibaware,
  title={VibAware: Context-Aware Tap and Swipe Gestures Using Bio-Acoustic Sensing},
  author={Kim, Jina and Kim, Minyung and Lee, Woo Suk and Yoon, Sang Ho},
  booktitle={Proceedings of the 2023 ACM Symposium on Spatial User Interaction},
  pages={1--12},
  year={2023}
}

@article{fan2022reducing,
  title={Reducing the Latency of Touch Tracking on Ad-Hoc Surfaces},
  author={Fan, Neil Xu and Xiao, Robert},
  journal={Proceedings of the ACM on Human-Computer Interaction},
  volume={6},
  number={ISS},
  pages={489--499},
  year={2022},
  publisher={ACM New York, NY, USA}
}

@inproceedings{pressurevision,
  title={PressureVision++: Estimating Fingertip Pressure from Diverse RGB Images},
  author={Grady, Patrick and Collins, Jeremy A and Tang, Chengcheng and Twigg, Christopher D and Aneja, Kunal and Hays, James and Kemp, Charles C},
  booktitle={Proceedings of the IEEE/CVF Winter Conference on Applications of Computer Vision},
  pages={8698--8708},
  year={2024}
}

@inproceedings{streli2022taptype,
  title={TapType: Ten-finger text entry on everyday surfaces via Bayesian inference},
  author={Streli, Paul and Jiang, Jiaxi and Fender, Andreas Rene and Meier, Manuel and Romat, Hugo and Holz, Christian},
  booktitle={Proceedings of the 2022 CHI Conference on Human Factors in Computing Systems},
  pages={1--16},
  year={2022}
}

@inproceedings{palmeira2023quantifying,
  title={Quantifying the'Gorilla Arm'Effect in a Virtual Reality Text Entry Task via Ray-Casting: A Preliminary Single-Subject Study},
  author={Palmeira, Eduardo GQ and Campos, Alexandre and Moraes, {\'I}gor A and de Siqueira, Alexandre G and Ferreira, Marcelo GG},
  booktitle={Proceedings of the 25th Symposium on Virtual and Augmented Reality},
  pages={274--278},
  year={2023}
}

@inproceedings{speicher2018selection,
  title={Selection-based text entry in virtual reality},
  author={Speicher, Marco and Feit, Anna Maria and Ziegler, Pascal and Kr{\"u}ger, Antonio},
  booktitle={Proceedings of the 2018 CHI conference on human factors in computing systems},
  pages={1--13},
  year={2018}
}

@article{bowman2012questioning,
  title={Questioning naturalism in 3D user interfaces},
  author={Bowman, Doug A and McMahan, Ryan P and Ragan, Eric D},
  journal={Communications of the ACM},
  volume={55},
  number={9},
  pages={78--88},
  year={2012},
  publisher={ACM New York, NY, USA}
}

@inproceedings{hansberger2017dispelling,
  title={Dispelling the gorilla arm syndrome: the viability of prolonged gesture interactions},
  author={Hansberger, Jeffrey T and Peng, Chao and Mathis, Shannon L and Areyur Shanthakumar, Vaidyanath and Meacham, Sarah C and Cao, Lizhou and Blakely, Victoria R},
  booktitle={Virtual, Augmented and Mixed Reality: 9th International Conference, VAMR 2017, Held as Part of HCI International 2017, Vancouver, BC, Canada, July 9-14, 2017, Proceedings 9},
  pages={505--520},
  year={2017},
  organization={Springer}
}

@inproceedings{potts2022tangibletouch,
  title={TangibleTouch: A Toolkit for Designing Surface-based Gestures for Tangible Interfaces},
  author={Potts, Dominic and Dabravalskis, Martynas and Houben, Steven},
  booktitle={Proceedings of the Sixteenth International Conference on Tangible, Embedded, and Embodied Interaction},
  pages={1--14},
  year={2022}
}

@inproceedings{xu2022enabling,
  title={Enabling hand gesture customization on wrist-worn devices},
  author={Xu, Xuhai and Gong, Jun and Brum, Carolina and Liang, Lilian and Suh, Bongsoo and Gupta, Shivam Kumar and Agarwal, Yash and Lindsey, Laurence and Kang, Runchang and Shahsavari, Behrooz and others},
  booktitle={Proceedings of the 2022 CHI Conference on Human Factors in Computing Systems},
  pages={1--19},
  year={2022}
}

@inproceedings{shadowtouch,
  title={Shadowtouch: Enabling free-form touch-based hand-to-surface interaction with wrist-mounted illuminant by shadow projection},
  author={Liang, Chen and Wang, Xutong and Li, Zisu and Hsia, Chi and Fan, Mingming and Yu, Chun and Shi, Yuanchun},
  booktitle={Proceedings of the 36th Annual ACM Symposium on User Interface Software and Technology},
  pages={1--14},
  year={2023}
}

@manual{OVRHand,
  title        = {OVRHand Class},
  author       = {Meta Platforms},
  organization = {Meta Platforms},
  year         = {2024},
  url          = {https://developers.meta.com/horizon/reference/unity/v72/class_o_v_r_hand},
  note         = {Accessed: 2025-04-07}
}

@misc{xreal2023,
  author       = {Xreal},
  title        = {Xreal},
  howpublished = {\url{https://www.xreal.com}},
}

@manual{unityarplane,
  title        = {Plane Detection},
  author       = {{Unity Technologies}},
  organization = {Unity Technologies},
  url          = {https://docs.unity3d.com/Packages/com.unity.xr.arfoundation@5.1/manual/features/plane-detection/plane-detection.html},
}

@inproceedings{zring,
  title={Z-ring: Single-point bio-impedance sensing for gesture, touch, object and user recognition},
  author={Waghmare, Anandghan and Ben Taleb, Youssef and Chatterjee, Ishan and Narendra, Arjun and Patel, Shwetak},
  booktitle={Proceedings of the 2023 CHI Conference on Human Factors in Computing Systems},
  pages={1--18},
  year={2023}
}

@inproceedings{touchinsight,
  title={TouchInsight: Uncertainty-aware Rapid Touch and Text Input for Mixed Reality from Egocentric Vision},
  author={Streli, Paul and Richardson, Mark and Botros, Fadi and Ma, Shugao and Wang, Robert and Holz, Christian},
  booktitle={Proceedings of the 37th Annual ACM Symposium on User Interface Software and Technology},
  pages={1--16},
  year={2024}
}

@inproceedings{structuredlightspeckle,
  title={Structured Light Speckle: Joint ego-centric depth estimation and low-latency contact detection via remote vibrometry},
  author={Streli, Paul and Jiang, Jiaxi and Rossie, Juliete and Holz, Christian},
  booktitle={Proceedings of the 36th Annual ACM Symposium on User Interface Software and Technology},
  pages={1--12},
  year={2023}
}

@inproceedings{halotouch,
  title={HaloTouch: Using IR Multi-Path Interference to Support Touch Interactions with General Surfaces},
  author={Xia, Ziyi and Huang, Xincheng and Fels, Sidney S and Xiao, Robert},
  booktitle={Proceedings of the 2025 CHI Conference on Human Factors in Computing Systems},
  pages={1--17},
  year={2025}
}

@inproceedings{scratchinput,
  title={Scratch input: creating large, inexpensive, unpowered and mobile finger input surfaces},
  author={Harrison, Chris and Hudson, Scott E},
  booktitle={Proceedings of the 21st annual ACM symposium on User interface software and technology},
  pages={205--208},
  year={2008}
}

@inproceedings{thermaltouch,
  title={Thermal touch: Thermography-enabled everywhere touch interfaces for mobile augmented reality applications},
  author={Kurz, Daniel},
  booktitle={2014 IEEE International Symposium on Mixed and Augmented Reality (ISMAR)},
  pages={9--16},
  year={2014},
  organization={IEEE}
}

@inproceedings{acustico,
  title={Acustico: Surface tap detection and localization using wrist-based acoustic TDOA sensing},
  author={Gong, Jun and Gupta, Aakar and Benko, Hrvoje},
  booktitle={Proceedings of the 33rd annual acm symposium on user interface software and technology},
  pages={406--419},
  year={2020}
}
